\documentclass{article}

\usepackage{microtype}
\usepackage{graphicx}
\usepackage{subfigure}
\usepackage{booktabs} 

\usepackage{hyperref}

\usepackage[accepted]{icml2025}

\usepackage{amsmath}
\usepackage{amssymb}
\usepackage{mathtools}
\usepackage{amsthm}

\usepackage[capitalize,noabbrev]{cleveref}

\theoremstyle{plain}

\theoremstyle{definition}

\theoremstyle{remark}

\usepackage[textsize=tiny]{todonotes}

\icmltitlerunning{NanoVLMs: How small can we go and still make coherent Vision Language Models?}

\begin{document}

\twocolumn[
\icmltitle{NanoVLMs: How small can we go and still make coherent Vision Language Models?}



\icmlsetsymbol{equal}{*}

\begin{icmlauthorlist}
\icmlauthor{Mukund Agarwalla*}{yyy}
\icmlauthor{Himanshu Kumar*}{yyy}
\icmlauthor{Raj Dandekar}{comp}
\icmlauthor{Rajat Dandekar}{comp}
\icmlauthor{Sreedath Panat}{comp}
\end{icmlauthorlist}

\icmlaffiliation{yyy}{Department of Computer Science, Indian Institute of Information Technology Nagpur, Maharashtra, India}
\icmlaffiliation{comp}{Vizuara Technologies, Pune, India}

\icmlcorrespondingauthor{Mukund Agarwalla}{mukundagarwalla2002@gmail.com}
\icmlcorrespondingauthor{Himanshu Kumar}{himanshukumariiitn@gmail.com}
\icmlcorrespondingauthor{Raj Dandekar}{raj@vizuara.com}

\icmlkeywords{Machine Learning, ICML, VLM, LLM, Rogue} 

\vskip 0.3in]



\printAffiliationsAndNotice{\icmlEqualContribution} 

\begin{abstract}
Vision-Language Models (VLMs), such as GPT-4V  and Llama 3.2 vision, have
garnered significant research attention for their ability to leverage Large
Language Models (LLMs) in multimodal tasks. However, their potential is
constrained by inherent challenges, including proprietary restrictions, substantial
computational demands, and limited accessibility. Smaller models, such as
GIT and BLIP, exhibit marked limitations, often failing to generate coherent and
consistent text beyond a few tokens, even with extensive training. This
underscores a pivotal inquiry: how small can a VLM be and still produce fluent and
consistent text? Drawing inspiration from the exceptional learning process of 3-4 year old children, who rely heavily on visual cues for understanding and
communication, we introduce two novel datasets: ShortDesc (featuring concise
image descriptions) and LongDesc (containing more detailed image descriptions).
These datasets consist of image-text pairs where the text is restricted to the simple
vocabulary and syntax typically used by young children, generated with a scaled-
down model, GPT-4o. Using these datasets, we demonstrate that it is possible to
train VLMs that are significantly smaller—up to 10 times smaller than state-of-the-
art(SOTA) small VLMs while maintaining architectural simplicity. To evaluate the
outputs, we leverage GPT-4o to grade the text, as if stories written by students, on
creativity, meaningfulness, and consistency, assigning scores out of 10. This
method addresses limitations of standard benchmarks by accommodating
unstructured outputs and providing a multidimensional evaluation of the model’s
capabilities. Our findings contribute to the development of lightweight, accessible
multimodal models for resource-constrained environments.
\end{abstract}
\begin{figure}[htb]
    \centering
    \includegraphics[width=1.0\linewidth]{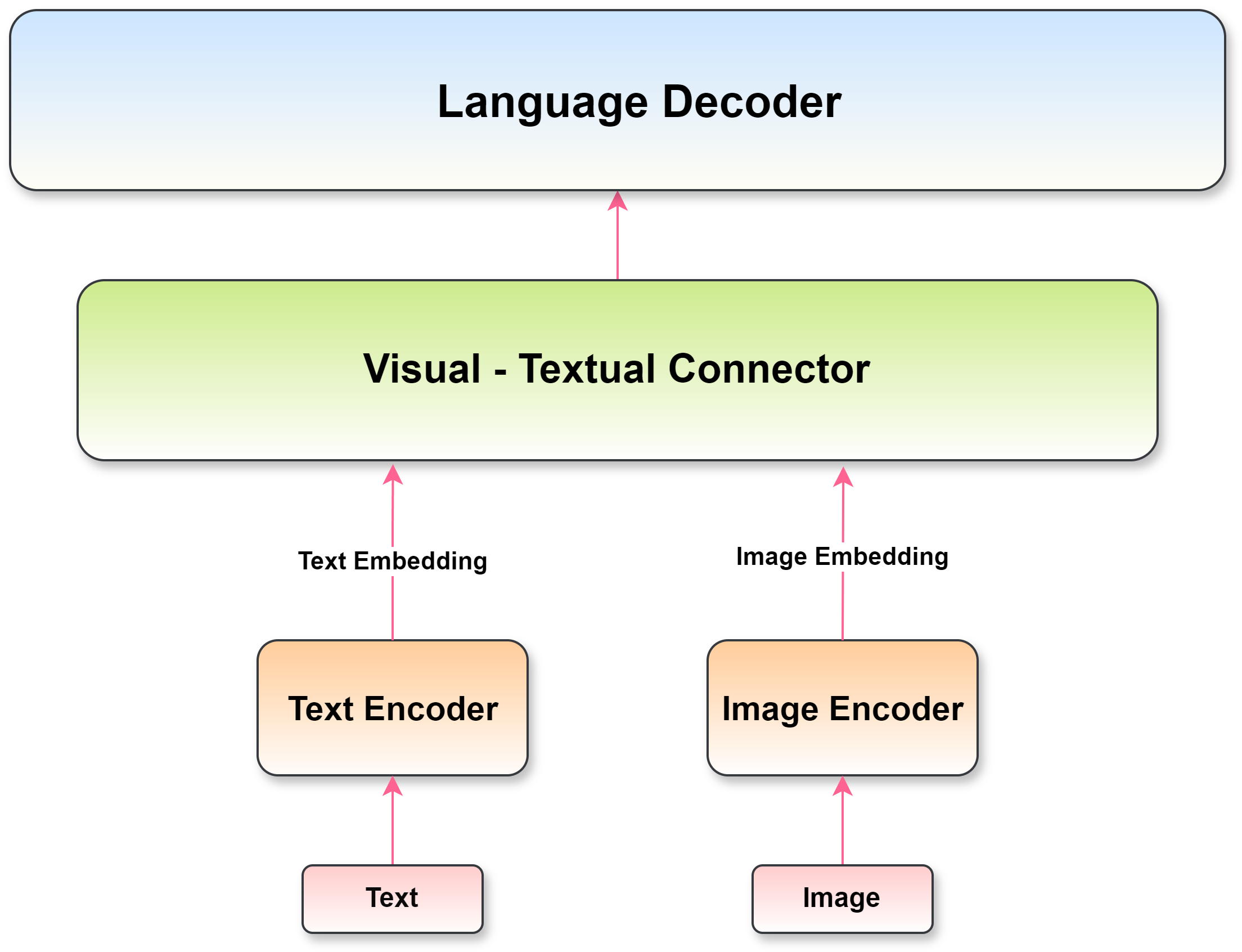}
    \caption{Root level architecture of VLM.}
    \label{fig: root level architecture of VLM's}
\end{figure}


\section{Introduction}
\label{submission}
LLMs\citep{zheng2023judgingllmasajudgemtbenchchatbot,zhao2024surveylargelanguagemodels, chatgpt2023, gpt4_2023} have significantly advanced natural language processing (NLP), demonstrating strong capabilities in  reasoning, long-form content generation and in-context learning (ICL). While models like GPT-3, LLaMA\cite{llama3_2024}, and Claude have achieved SOTA performance across text-based tasks, their unimodal nature limits their ability to process and interpret visual data. The rise of VLMs has bridged this gap by integrating pretrained vision encoders with large-scale language models, enabling multimodal reasoning across tasks such as Image Captioning (IC)\citep{vinyals2015show}, Visual Question Answering (VQA)\citep{antol2015vqa}, Optical Character Recognition (OCR)\citep{shi2017ocr}, and Visual Grounding\citep{plummer2015visualgrounding}.
\begin{figure}[htb]
    \centering
    \includegraphics[width=\linewidth]{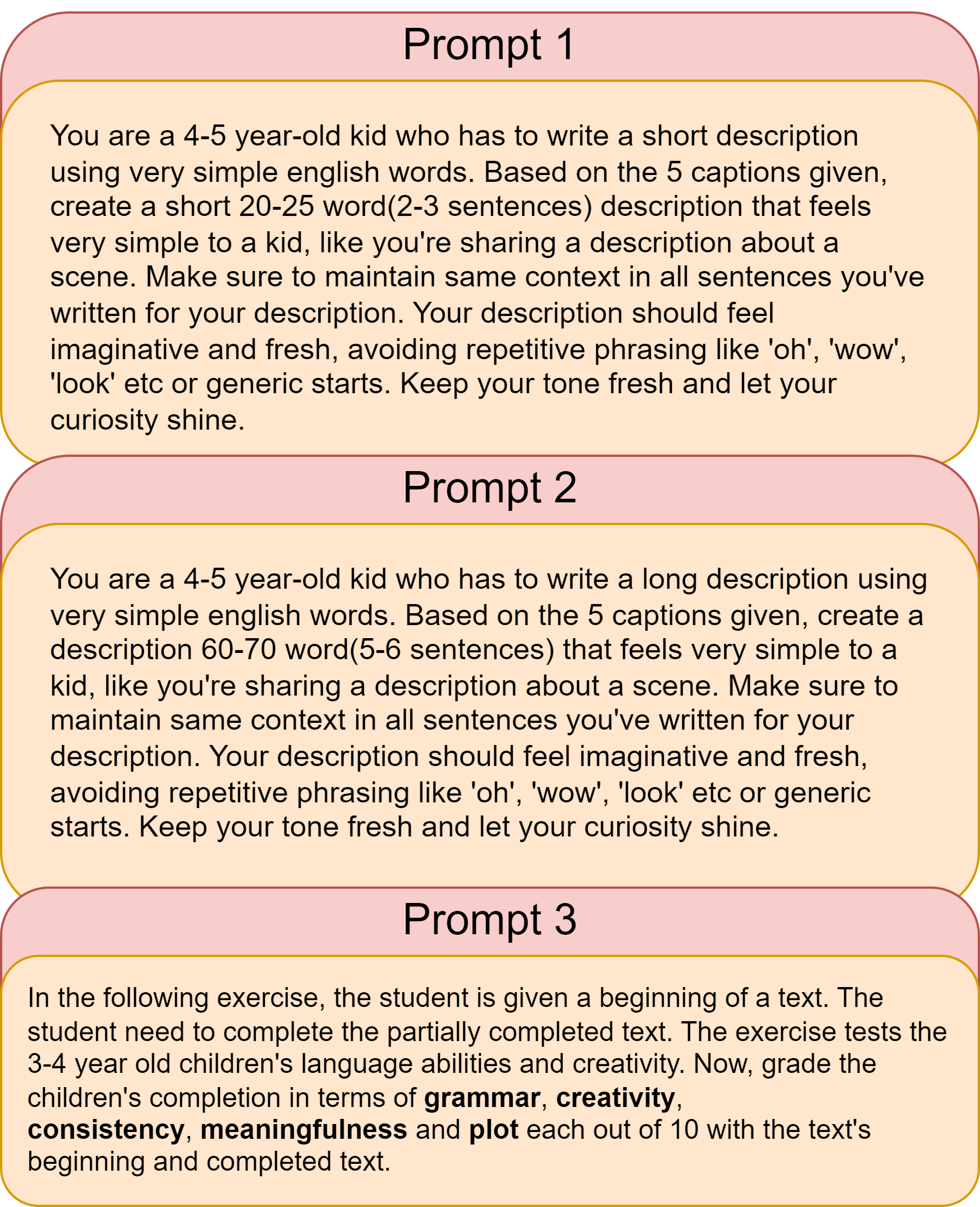}
    \caption{Prompts to GPT-4o for dataset creation and evaluation.}
    \label{fig:prompts}
\end{figure}
\label{submission2}
Recent breakthroughs in multimodal learning\citep{akkus2023multimodaldeeplearning} have led to the development of high-performing VLMs such as PaLI-X\citep{luo2024vividmedvisionlanguagemodel}, GPT-4V\citep{ghosh2024exploringfrontiervisionlanguagemodels}, LLaVA\citep{liu2023visualinstructiontuning}, Flamingo\citep{alayrac2022flamingovisuallanguagemodel}, Qwen2.5-VL-7B-Instruct\citep{wang2024qwen2vlenhancingvisionlanguagemodels}, MiniGPT-4\cite{zhu2024minigpt}, and InstructBLIP\cite{ghosh2024exploringfrontiervisionlanguagemodels}. These models typically comprise three core architectural components: (1) a visual encoder\citep{kar2024brave,jain2023vcoder}, responsible for transforming raw images into feature-rich representations using pretrained models like CLIP\citep{radford2021learningtransferablevisualmodels}, ViT\cite{ghosh2024exploringfrontiervisionlanguagemodels} or BEiT\citep{ghosh2024exploringfrontiervisionlanguagemodels}; (2) a visual-textual connector\citep{cha2024honeybeelocalityenhancedprojectormultimodal,huggingface2023vlp}, which aligns and fuses multimodal features through cross-attention mechanisms or learned projection layers; and (3) a language decoder, often based on autoregressive transformers like LLaMA, Gemma, Phi-3 or GPT, which generates coherent textual outputs grounded in visual context. Hybrid approaches such as SPHINX\citep{ahuja2024sphinxsampleefficientmultilingual} and mPLUG-Owl\cite{ye2024mplugowlmodularizationempowerslarge} have explored different fusion strategies to improve cross-modal understanding.

However, scaling VLMs to billions of parameters presents computational and memory constraints, limiting their accessibility for researchers and real-world applications. While compact models such as SmolVLM-256M, TinyGPT-V\citep{yuan2024tinygptvefficientmultimodallarge}, BLIP-base\citep{li2022blipbootstrappinglanguageimagepretraining},OFA-Tiny, GIT\citep{wang2022gitgenerativeimagetotexttransformer}, and Kosmos-2\citep{peng2023kosmos2groundingmultimodallarge} have demonstrated promising efficiency, they often struggle with fine-grained visual reasoning and multimodal consistency. In this work, we introduce NanoVLMs, a family lightweight yet effective VLM that optimizes parameter allocation across the three core components, prioritizing efficient visual encoding and refined cross-modal alignment. Beyond efficiency, we analyze their real-world applicability by evaluating how well they retain semantic accuracy, adaptability to varying input complexities, and robustness in handling diverse visual-textual associations.

\begin{figure*}[htb]
    \centering
    \includegraphics[width=\textwidth]{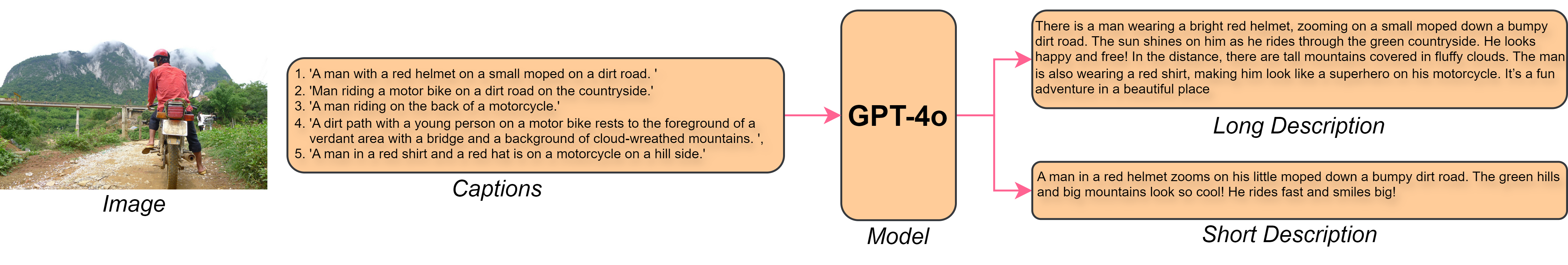}
    \caption{Process for creation of datasets.}
    \label{fig:dataset creation}
\end{figure*}

\section{Methodology}
\label{submission3}
This section outlines the methodologies employed to develop NanoVLMs, a highly efficient VLM that is more than 10 times smaller than SOTA VLMs while retaining competitive performance. To the best of our knowledge, NanoVLMs are the first to achieve such extreme compression without compromising on results. To achieve this, the data used for model training should be simple and easy to understand. aligning with the cognitive abilities of 3–4-year old children. Inspired by the learning processes of 3-4 year old children, we adopted an analogy based on how children in this age group acquire knowledge. A similar analogy was explored in  \citep{eldan2023tinystoriessmalllanguagemodels}, which focused exclusively on textual data. However, we found this approach inadequate, as 3–4 year old children are more reliant on visual stimuli than text for learning. Recognizing this, we designed and trained a minimalist VLM with a straightforward architecture, leveraging simple images and text that could be easily understood by children. 

The high-level architectural design and training process draw inspiration from methodologies employed in SOTA VLMs while emphasizing efficiency and scalability. Our work is structured around several key components: a robust dataset preparation framework that ensures diverse and high-quality training data, the NanoVLMs architecture designed to optimize parameter efficiency without sacrificing performance, a comprehensive training and experimentation pipeline tailored for effective convergence, and a rigorous evaluation methodology that examines both quantitative and qualitative aspects of generated descriptions. We also implement techniques to enhance cross-modal interactions, ensuring that the model effectively captures fine-grained visual details.

\subsection{Dataset}
\subsubsection{Dataset Overview }
To train NanoVLMs, we utilized the COCO (Common Objects in Context)\citep{lin2014microsoft,chen2015microsoft} dataset, a widely recognized resource in computer vision tasks. This dataset is ideal for our study as it features high-resolution, richly annotated images from diverse domains, including people, animals, food, vehicles, and outdoor settings—perfectly aligning with the learning analogy of 3–4 year old children. For our work, we specifically leveraged the image-captioning component of COCO, where each image is paired with five natural language captions describing the scene and its objects. From this dataset, we selected approximately 28K image-caption pairs, using 90 percent for training and 10 percent for validation. Additionally, to evaluate the model’s knowledge, versatility, and generalization capabilities, we tested it on 25 separate data samples that were entirely distinct from the training and validation sets.  

\subsubsection{Dataset Preparation }
To prepare the dataset, we used the COCO dataset's images and corresponding captions to generate image descriptions, constructing two datasets: ShortDesc and LongDesc. Specifically, ShortDesc comprises concise image descriptions of 20–25 words, while LongDesc features detailed image descriptions of 60–70 words. These datasets were designed to assess how the model handles shorter versus longer text inputs, reflecting its ability to process and generate meaningful and consistent outputs. This mirrors the developmental process of 3–4 year old children, who acquire intellectual abilities through exposure to diverse visuals along with various linguistic patterns. 

For generating these descriptions, we employed OpenAI’s GPT-4o, a SOTA text generation model capable of producing high-quality synthetic content. Combined captions for each image (all captions) along with the respective prompt is passed to GPT-4o\citep{openai2024gpt4o}, where Prompt1 shown in Figure 2 is used to generate ShortDesc dataset and Prompt2 shown in Figure 2 is used to generate LongDesc dataset. The model produced outputs based on the structure and constraints defined in the respective prompts. Figure 3 illustrates the process of prompt passing and dataset preparation.

\subsection{Architecture}
The primary objective of NanoVLMs is to complete partially provided textual descriptions by generating coherent and contextually appropriate outputs. To achieve this, we designed a VLM with a simple yet effective transformer-based architecture consisting of three key components: a visual encoder for processing images, a visual-textual connector to bridge visual and textual modalities, and a language decoder for generating text as shown in Figure 1.
\begin{figure*}[htb]
    \centering
    \includegraphics[width=\textwidth]{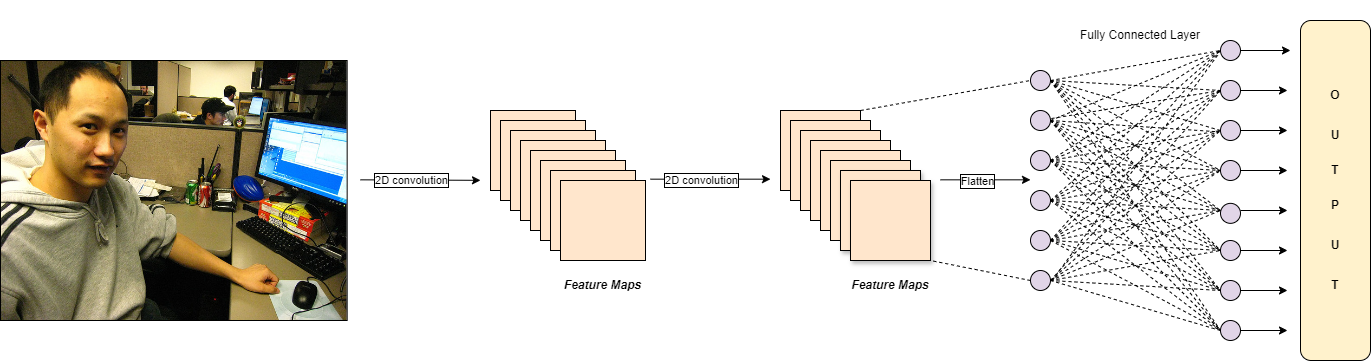}
    \caption{Feature extraction from an image.}
    \label{fig:feature extraction}
\end{figure*}
\begin{figure*}[htb]
    \centering
    \includegraphics[width=\textwidth]{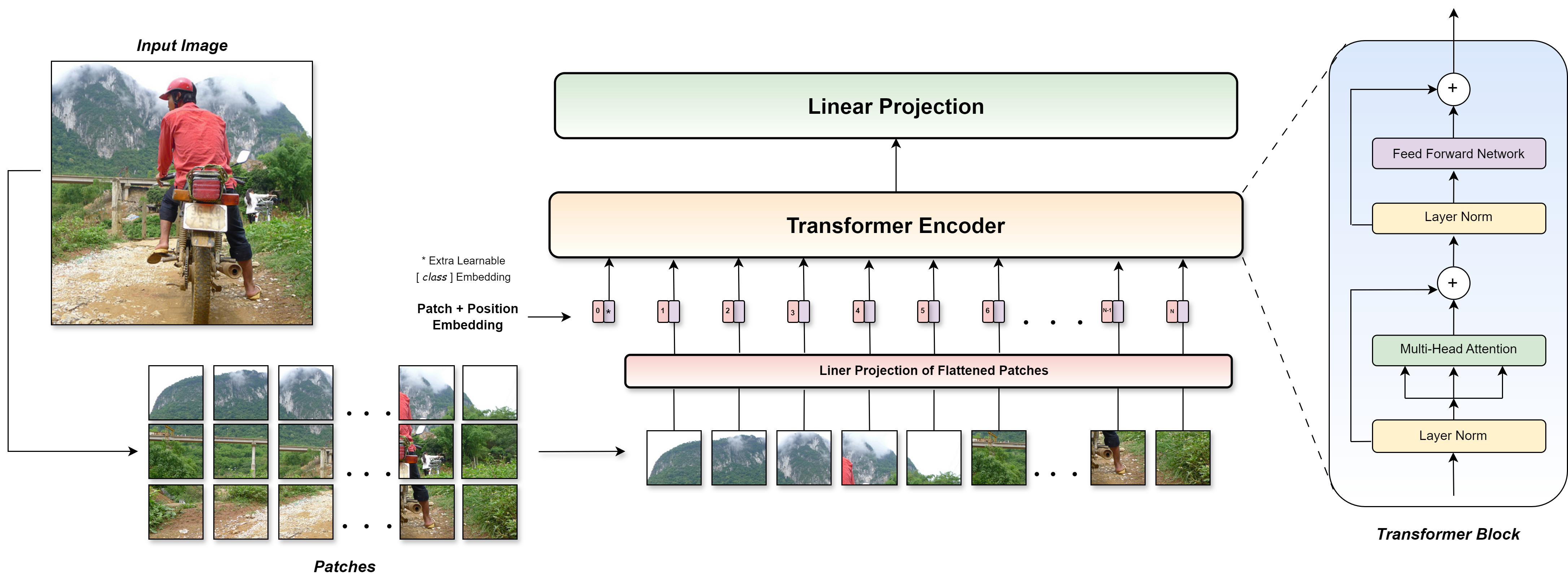}
    \caption{Vision Transformer}
    \label{fig:vision transformer}
\end{figure*}

The core of the NanoVLM architecture lies in its transformer blocks (shown in Figure 5), which form the foundation of both the visual encoder and the language decoder. Each transformer block comprises multi-head attention\citep{jalammar2019gpt2,vaswani2023attentionneed} for capturing relationships across input tokens—whether image patches or text—and a multi-layer perceptron (MLP) for processing the outputs of the attention mechanism. To ensure stable training and faster convergence, layer normalization is applied prior to the attention and MLP layers. A key distinction in the decoder is the use of causal self-attention, where masking is employed to uphold the autoregressive nature of text generation. This mechanism is vital for maintaining coherence and contextual accuracy, ensuring that predictions are based solely on prior information, a critical requirement for generating fluent and logically consistent textual descriptions. 

\subsubsection{Visual Encoder }
The visual encoder in NanoVLM is a critical component responsible for extracting meaningful features from images, drawing inspiration from the Vision Transformer (ViT) architecture while being optimized for compactness. To maintain performance, we process images at a resolution of 224x224 pixels\citep{thapa2024dragonflymultiresolutionzoominencoding}, dividing them into 16x16 pixel patches to yield 196\citep{wen2024efficientvisionlanguagemodelssummarizing} patches per image. These patches undergo a series of transformations beginning with patch embedding, where the image is passed through two 2D convolutional layers(as shown in Figure 4) followed by layer normalization\citep{ba2016layernormalization} and ReLU\citep{agarap2019deeplearningusingrectified} activation. This is succeeded by a fully connected neural network, which transforms the patches into 196 tokens. A [CLS] token is then prepended, making the sequence 197 tokens. Positional encoding is applied to retain spatial information, followed by normalization. These enriched embeddings are then processed through a series of transformer blocks, where multi-head attention mechanisms capture contextual dependencies between the patches. Finally, the [CLS] token is aggregated to form a compact representation that encapsulates the salient features of the image. This streamlined yet robust approach ensures effective visual feature extraction while keeping the model size minimal. 
\begin{table}[t]
\caption{Variable hyperparameters of NanoVLMs.}
\vskip 0.15in
\begin{center}
\begin{small}
\begin{tabular}{lccc}
\toprule
Parameters & \textbf{mini} & \textbf{base} & \textbf{large} \\
\midrule
n\_blks & 1 & 3 & 5 \\
n\_layer & 4 & 8 & 10 \\
n\_head & 8 & 8 & 16 \\
head\_size & 12 & 16 & 12 \\
n\_embd & 96 & 128 & 192 \\
img\_embd\_dim & 400 & 512 & 512 \\
\midrule
\textbf{Total parameters} & 5M & 16M & 25M \\
\bottomrule
\end{tabular}
\end{small}
\end{center}
\vskip -10mm
\end{table}

\subsubsection{ Visual-Textual Connector }
The visual-textual connector is a pivotal component in the NanoVLMs architecture, responsible for bridging the gap between the visual and textual modalities. The visual embeddings and the textual embeddings must be aligned in the same dimensional space to enable effective interaction between the two modalities. To achieve this, we employ a multimodal projector that consists of a single learnable layer followed by GELU that reduces the dimensionality of the visual embeddings. Once the visual embeddings are projected into the textual embedding space, both the visual and textual embeddings are concatenated to form a multimodal token embedding. This combined representation effectively encapsulates both the image’s content and its corresponding textual description. The resulting multimodal token embedding is then passed as input to the decoder block, where it will guide the generation of coherent and contextually relevant textual descriptions.  

\begin{figure*}[htb]
    \centering
    \includegraphics[width=\textwidth]{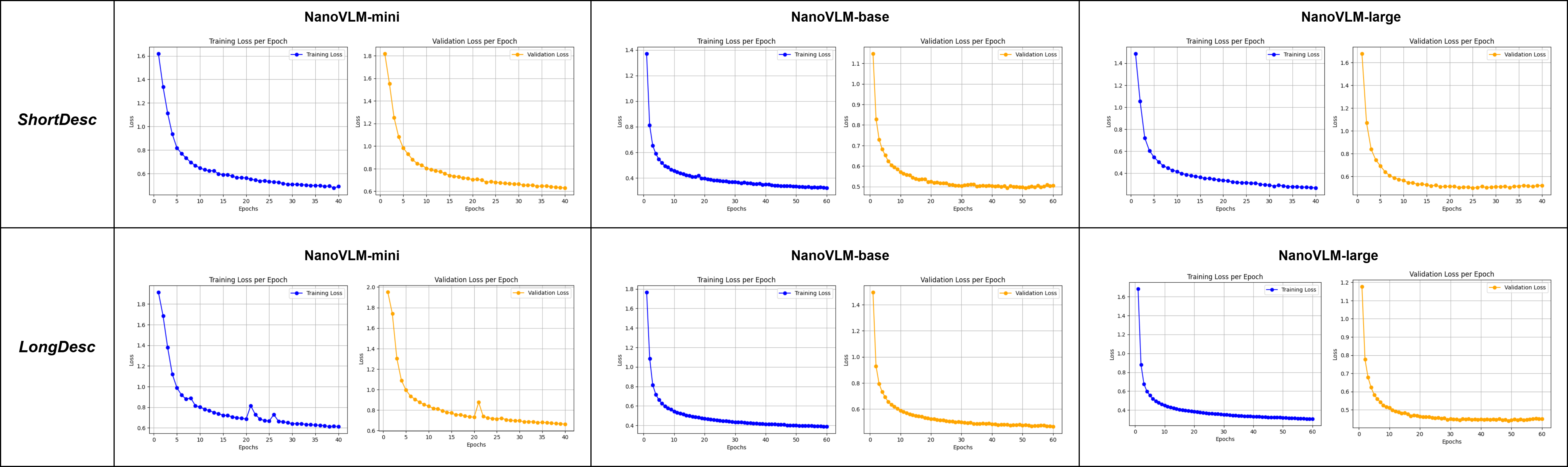}
    \caption{Training and validation losses of NanoVLMs.}
    \label{fig:loss graphs}
\end{figure*}

\subsubsection{Decoder Block}
The decoder block in NanoVLM transforms fused visual-textual embeddings into coherent text using a transformer-based architecture, ensuring text generation. It begins by passing the multimodal token embedding through a positional embedding layer, which encodes token order. The input then moves through transformer blocks with multi-head self-attention, but unlike the encoder, the decoder applies causal self-attention, masking\citep{liu2022forgetful,yin2024stablemask} future tokens to prevent information leakage and enforce autoregressive generation. Finally, the processed output undergoes layer normalization and a linear projection, mapping it to a vocabulary space where logits determine the next token. This structured decoding mechanism enables NanoVLM to generate fluent, context-aware descriptions when provided with both an image and partial text as input.We employ cross-entropy loss to compute the error between the predicted and actual target text. This loss is used to guide the training of the model, optimizing the parameters to generate accurate and coherent textual descriptions.

\subsection{Experiments}
This section details the experimental setup and hyperparameter tuning for training all three versions of NanoVLM. The models were trained on a single A100 GPU, with key hyperparameters such as \textbf{n\_blks} (number of transformer blocks in the visual encoder), \textbf{n\_layer} (number of transformer layers in decoder), \textbf{n\_head} (number of attention heads), \textbf{head\_size} (size of each head), \textbf{n\_embd} (textual embedding dimension), and \textbf{img\_embd\_dim} (visual embedding dimension) gradually scaled up as we moved from the mini to the large version, as shown in Table 1. This progressive scaling allowed us to analyze how increasing model capacity influenced performance while maintaining computational efficiency. Certain hyperparameters remained fixed across all versions to ensure stability during training, including dropout = 0.1, image\_size = 224x224, patch\_size = 16x16, and learning rate = 1e-3. Additionally, we present the distribution of total learnable parameters for each version of the model across the three core modules in Table 2. Since 3-4 year old children primarily learn through visual cues, we allocated a larger portion of the model’s parameters to the visual encoder module, ensuring that the extracted features from images were rich and informative while maintaining an efficient balance between vision and language processing. 
\begingroup

\begin{table}[t]
\caption{Distribution of number of parameters across different modules in NanoVLMs.}
\vskip 0.15in
\begin{center}
\begin{small}
\begin{tabular}{lccc}
\toprule 
Module & \textbf{mini} & \textbf{base} & \textbf{large} \\
\midrule
Visual Encoder & \textbf{69\%} & \textbf{78\%} & \textbf{73\%} \\
Multimodal Projector & 14\% & 8\% & 6\% \\
Decoder & 17\% & 16\% & 21\% \\

\bottomrule
\end{tabular}
\end{small}
\end{center}
\vskip -0.2in
\end{table}

\endgroup
\begin{figure*}[htb]
    \centering
    \includegraphics[width=\textwidth]{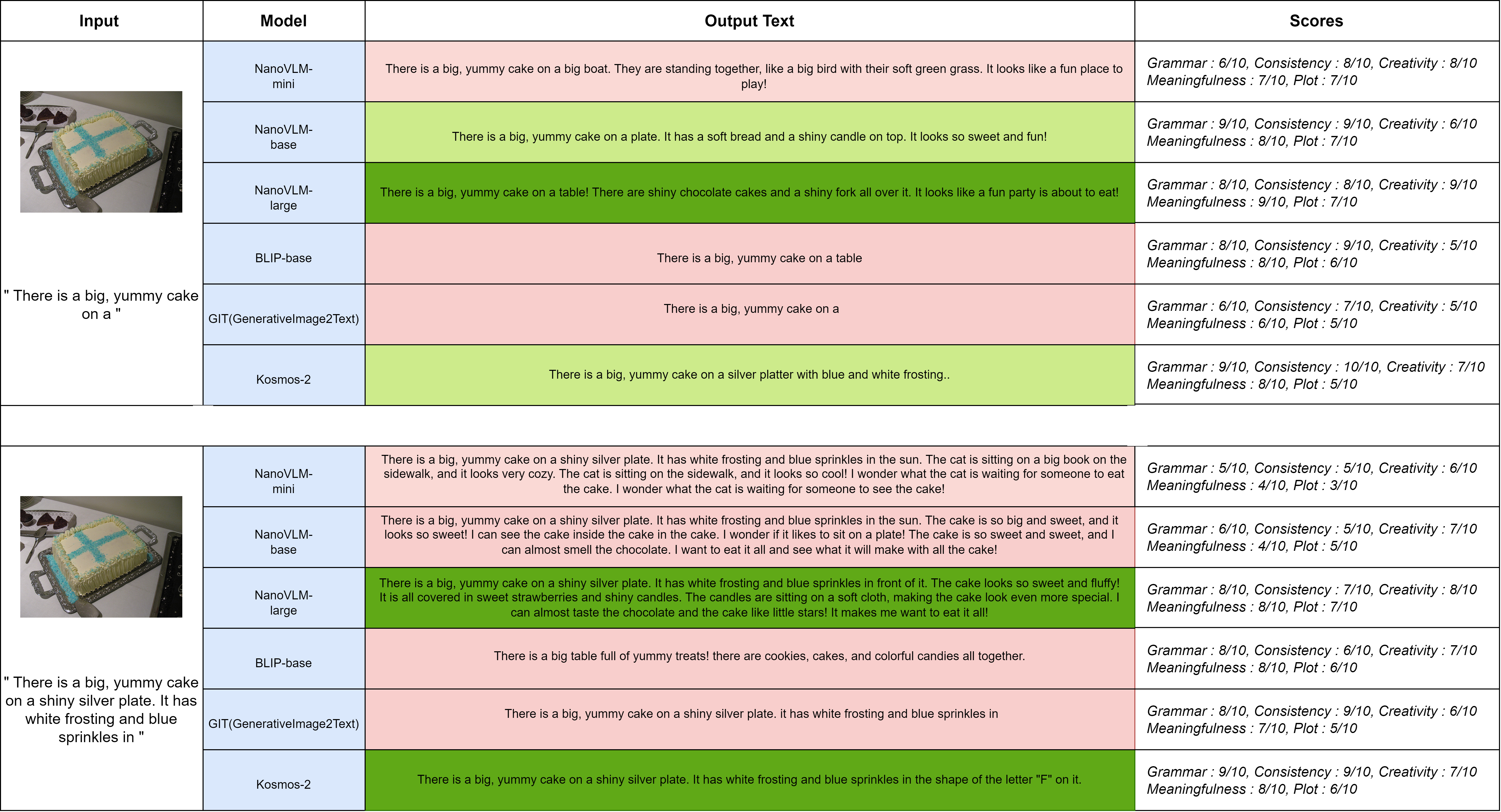}
    \caption{Sample output text and  evaluation scores of various models on short and long partial text completion task.}
    \label{fig:Sample outputs on ShortDesc}
\end{figure*}

\section{Evaluation}
Traditional evaluation of VLMs typically relies on structured benchmark datasets where the model’s output is compared against a predefined ground-truth answer. To comprehensively evaluate a VLM, we focus on five key benchmarks—grammatical correctness, consistency, creativity, meaningfulness, and plot—each of which plays a crucial role in determining the model’s ability to generate structured and engaging descriptions. Our primary objective is to investigate whether a VLM with as few as 6M–25M parameters can still generate coherent and contextually relevant text. 
Inspired by the evaluation framework of \citep{eldan2023tinystoriessmalllanguagemodels}, we employ an LLM-based evaluation approach that leverages GPT-4o to assess generated text quality. Our evaluation setup consists of a manually curated dataset of 25 image descriptions, where each description's beginning along with its corresponding image, is provided as a prompt to NanoVLMs. The model then completes the partial text while attending to the image, and its output is subsequently graded using Prompt 3 (shown in Figure 2) by GPT-4o based on key evaluation benchmarks outlined in Table 3.
To ensure that our image description generation task is non-trivial, we deliberately structure the input prompts of 6–7 words long for short descriptions and 18–20 words long for long descriptions. This approach challenges the model’s ability to produce semantically meaningful and grammatically sound completions, especially when required to infer missing context from the provided image.
Furthermore, to verify that the model does not simply memorize training data, we conduct an analysis using ROUGE scores which is detailed in the section 4. 
By integrating these methods, we provide a comprehensive assessment of NanoVLMs' linguistic and contextual competence, addressing the limitations of traditional benchmark-driven approaches.

\begin{table*}[t]
\caption{Model comparison based on key metrics and model size.}
\vskip 0.15in
\begin{center}
\begin{small}
\begin{tabular}{ll cccccccc}
\toprule
\multicolumn{2}{c}{Dataset} & Model &  Size &Grammar & Creativity & Consistency & Meaningfulness & Plot & Average Total \\
\midrule
\multicolumn{2}{c}{\textbf{ShortDesc}}  & mini&5M   & 6.36 & 7.44 & 6.28 & 6.12 & 5.72 & 31.92 \\
           & & base &16M & 8.36 & 8.00 & 8.20 & 7.68& 6.56 & 38.80 \\
           & & large &25M& 8.20 & \textbf{8.00} & 7.83 & 7.79 & \textbf{7.75} & 39.39 \\
           & & BLIP-base &223M& 7.88 & 5.08 & 8.52 & 7.16 & 4.36 & 33.00 \\
           & & GIT &350M& 6.31 & 3.18 & 7.90 & 4.22 & 2.81 & 24.42 \\
           & & Kosmos-2 &1.3B& \textbf{8.73} & 7.15 & \textbf{9.15} & \textbf{8.15} & 6.68 & \textbf{39.86} \\
\midrule
\multicolumn{2}{c}{\textbf{LongDesc}}   & mini &5M& 5.92 & 7.36 & 5.36 & 5.72 & 4.56 & 28.92 \\
           & & base &16M & 7.56 & 8.32 & 7.40 & 7.40& 6.76 & 37.44 \\
           & & large &25M& 8.24 & \textbf{8.52} & 7.84 & 8.08 & \textbf{7.16} & 39.84 \\
           & & BLIP-base &223M& 7.28 & 5.36 & 8.62 & 6.41 & 4.50 & 32.17 \\
            & & GIT &350M& 6.83 & 4.54 & 7.62 & 5.75 & 4.04 & 28.78 \\
           & & Kosmos-2 &1.3B& \textbf{8.90} & 7.70 & \textbf{8.95} & \textbf{8.30} & 6.80 & \textbf{40.65} \\
\bottomrule
\end{tabular}
\end{small}
\end{center}
\vskip -0.1in
\end{table*}

\section{Results}
This section evaluates the effectiveness of NanoVLMs in generating accurate and coherent textual descriptions. We first analyze the training and validation losses to assess the model’s convergence and stability. Next, we evaluate the generated outputs on structured benchmarks, highlighting their relevance and fluency. Additionally, we perform a ROUGE score analysis, verifying the diversity and contextual alignment of the generated descriptions.

The training and validation loss trajectories, as illustrated in Figure 6, exhibit a consistent downward trend across all NanoVLM versions, affirming stable convergence and effective optimization. The loss curves reveal that the gap between training and validation losses remains minimal for NanoVLMs, with a maximum observed difference of only 0.08 to 0.1. However, NanoVLMs trained on LongDesc show slightly less stable convergence, primarily due to the increased complexity and length of textual descriptions. Generating a long and contextually rich description while maintaining consistency and meaningfulness is a challenging task for a compact model with limited parameters. Despite this, the loss curves across all versions eventually plateau in the final training epochs, demonstrating that the models successfully learn structured vision-language representations while mitigating overfitting. 

To rigorously evaluate the performance of our NanoVLM models, we compare their generated outputs, conditioned on visual input and partial text against three significantly larger VLMs: BLIP-base (223M), GIT (350M), and Kosmos-2 (1.3B). These models contain approximately 10x, 14x, and 50x times more trainable parameters than NanoVLM-large, respectively. Figure 7 presents a qualitative comparison between NanoVLMs and these large SOTA VLMs, analyzing their performance in the image description generation task. This figure specifically examines how each model handles both short and long partial text inputs, highlighting their ability to generate coherent and contextually relevant descriptions.

The generated descriptions using short partial text (upper section of Figure 7) highlight that NanoVLM-mini produces a more descriptive and creative output than BLIP-base and GIT, which struggle to generate meaningful text despite having significantly more parameters. NanoVLM-mini attempts to capture the image’s surroundings, though it occasionally misinterprets objects. While NanoVLM-mini emphasizes creativity, NanoVLM-base strikes a better balance between contextual accuracy and coherence. However, Kosmos-2, with its significantly larger parameter size, delivers a more consistent, meaningful, and concise description, outperforming both NanoVLM-mini and NanoVLM-base in creativity and contextual depth. NanoVLM-large, despite having 50x fewer parameters than Kosmos-2, achieves comparable scores in grammar, consistency, and meaningfulness. It excels in capturing surrounding elements with precision and demonstrates strong contextual understanding.

The lower section of the Figure 7 illustrates how our NanoVLMs perform on long text completion task. Unlike the results in Figure 7(upper section), NanoVLM-mini struggles significantly in maintaining coherence and relevance, generating descriptions that veer off-topic and introduce unrelated elements. This is expected, as smaller models often face difficulties in handling longer texts, leading to a loss of contextual grounding. In contrast, NanoVLM-base performs notably better, capturing more relevant details about the image, but it still exhibits some degree of repetition, inconsistency and capturing surroundings, suggesting limitations in long-text completion. Interestingly, BLIP-base and GIT also fails even worse to generate rich and contextually deep descriptions, with BLIP producing a more generic output that does not focus on the given input details, and GIT barely extending beyond the input partial text. NanoVLM-large, however, demonstrates a strong ability to generate structured and meaningful long descriptions. It not only maintains contextual relevance but also enriches the scene with well-placed details and identify the surrounding objects. The description is highly coherent and captures a level of depth that is missing in the smaller versions and competing models. Most notably, NanoVLM-large performs comparably to Kosmos-2, showcasing its efficiency in long-text completion despite its compact size. Kosmos-2 still holds a slight edge in capturing the minute details from the image, but NanoVLM-large proves to be a strong model for providing comprehensive description, excelling in both descriptive richness and fluency. 
\begin{figure*}[htb]
    \centering
    \includegraphics[width=\textwidth]{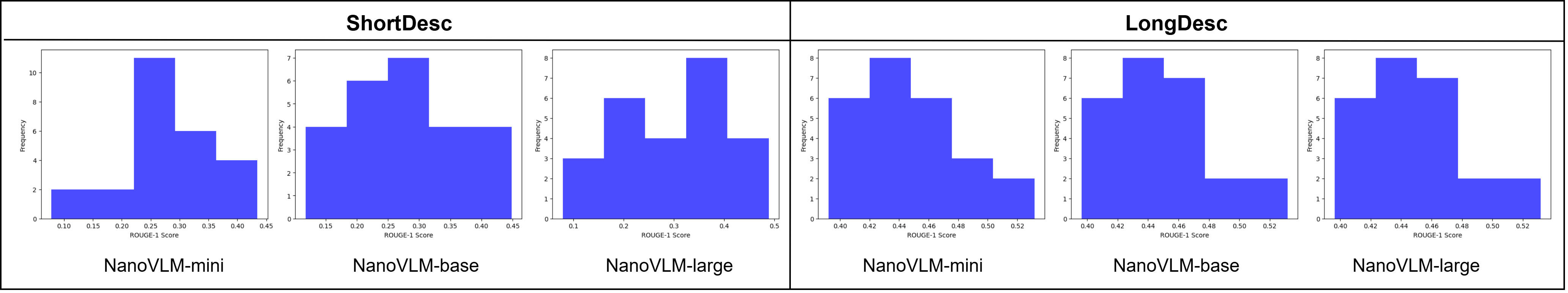}
    \caption{Histogram plot of rouge scores across each model.}
    \label{fig:histogram}
\end{figure*}

Table 3 presents a comprehensive evaluation of our NanoVLM models alongside three significantly larger SOTA VLMs, analyzing their performance on both short and long image descriptions. The values in this table represent the average scores(for 25 datapoints) calculated using the score assigned by GPT-4o for each benchmark. The models are assessed based on these average scores, culminating in an overall average score that reflects their ability to generate coherent textual outputs. By comparing these scores, we aim to highlight how our NanoVLM models, deliver competitive results while demonstrating distinct strengths across different aspects of language generation.

In the short text completion task, NanoVLM-base and NanoVLM-large consistently outperform BLIP-base and GIT across most benchmarks, particularly in creativity and plot, where they achieve higher scores compared to BLIP-base and GIT. NanoVLM-large also surpasses Kosmos-2 in creativity (8.00 vs. 7.15) and plot (7.75 vs. 6.68). While Kosmos-2 remains the strongest overall with a total score of 39.86, NanoVLM-large, with a total score of 39.39, is nearly comparable despite being 50 times smaller in parameter count. Additionally, BLIP-base and GIT exhibit highly uneven performance, excelling in consistency but significantly lacking in creativity, meaningfulness and plot. In contrast, our three NanoVLM models maintain a more balanced performance across all benchmarks, ensuring better coherence and overall stability in text completion.

For long text completion, a similar trend is observed, with NanoVLM-base and NanoVLM-large outperforming BLIP-base and GIT across most metrics. NanoVLM-large achieves 8.52 in creativity and 7.16 in plot, significantly surpassing BLIP-base, GIT, and even Kosmos-2. NanoVLM-large's performance in key metrics like grammar and meaningfulness is comparable to Kosmos-2. Despite its compact size, NanoVLM-large is nearly on par with Kosmos-2, demonstrating its efficiency in generating extended descriptions. Notably, our models maintain stable and well-distributed performance across all benchmarks, unlike BLIP-base and GIT, which struggle in areas like creativity and meaningfulness, resulting in less coherent and engaging narratives. These results highlight the potential of our NanoVLM family as an effective, lightweight alternative to larger VLMs, offering competitive performance while requiring significantly fewer resources.


While our NanoVLMs generate fluent and coherent English descriptions, their effectiveness would be undermined if they merely copied or paraphrased large portions of the training dataset. To demonstrate the originality of our models, we assess the diversity of their outputs by measuring word and n-gram overlap. Specifically, we compute the Rouge-1 score by comparing NanoVLM-completed descriptions with the actual descriptions from the training set. As shown in Figure 8, the Rouge-1 scores for all NanoVLM variants are remarkably low(approximately less than 0.5), confirming that our models produce highly diverse descriptions rather than memorizing or reusing training data. This low overlap is particularly beneficial, as it ensures that NanoVLMs generate original, contextually relevant descriptions rather than relying on simple recall, reinforcing their ability to generalize effectively while maintaining fluency and coherence.

\section{Conclusion}
Our introduction of NanoVLMs, a highly efficient family of VLMs built from scratch, places a strong emphasis on minimizing parameters while preserving performance. By systematically optimizing each component (encoder, decoder, and connector), we develop NanoVLM variants that are significantly smaller than conventional VLMs, raising fundamental questions about the training process and data requirements for building such models. To explore these aspects, we employ LongDesc and ShortDesc, highlighting the need for a more complex architecture to generate extended narratives compared to concise ones.

While our findings demonstrate that even with a small dataset, a well-designed, small-scale VLM can achieve competitive results, certain limitations remain. Increasing the dataset size could further enhance the model’s generalization capabilities, particularly for long-form descriptions. Additionally, the model's ability to generalize to more complex domains or handle fine-grained visual reasoning requires further exploration. Optimizing for compactness may also introduce trade-offs in multimodal alignment, potentially impacting performance on tasks requiring deep semantic understanding. Finally, although our evaluation provides structured insights into model performance, a more extensive human evaluation could help assess fluency, coherence, and real-world applicability.
This work lays the foundation for developing ultra-compact yet effective VLMs, making them more practical, accessible, and adaptable to real-world applications.



\clearpage 


\bibliography{example_paper}
\bibliographystyle{icml2025}

\clearpage  
\appendix
\onecolumn  
\section*{Appendix}
To provide a more comprehensive evaluation of our NanoVLM models, we extend the qualitative comparison presented in Figure 7 of the main paper by including five additional examples. These examples, presented in Figures 9, 10, 11, 12, and 13, cover a diverse range of visual inputs and partial text prompts, offering deeper insights into the strengths and limitations of each model. Each figure follows the same structure as Figure 7, with an upper section focusing on short and a lower section focusing on long, description completion using short and long partial input text respectively along with input image. This extended comparison allows for a clearer understanding of how NanoVLM-mini, NanoVLM-base, and NanoVLM-large perform relative to larger VLMs (BLIP-base, GIT, and Kosmos-2) in various scenarios. The results illustrate differences in coherence, contextual relevance, descriptive richness, and fluency, further validating the efficiency of NanoVLM models in image description tasks.

\begin{figure*}[htb]  
    \centering
    \includegraphics[width=\textwidth]{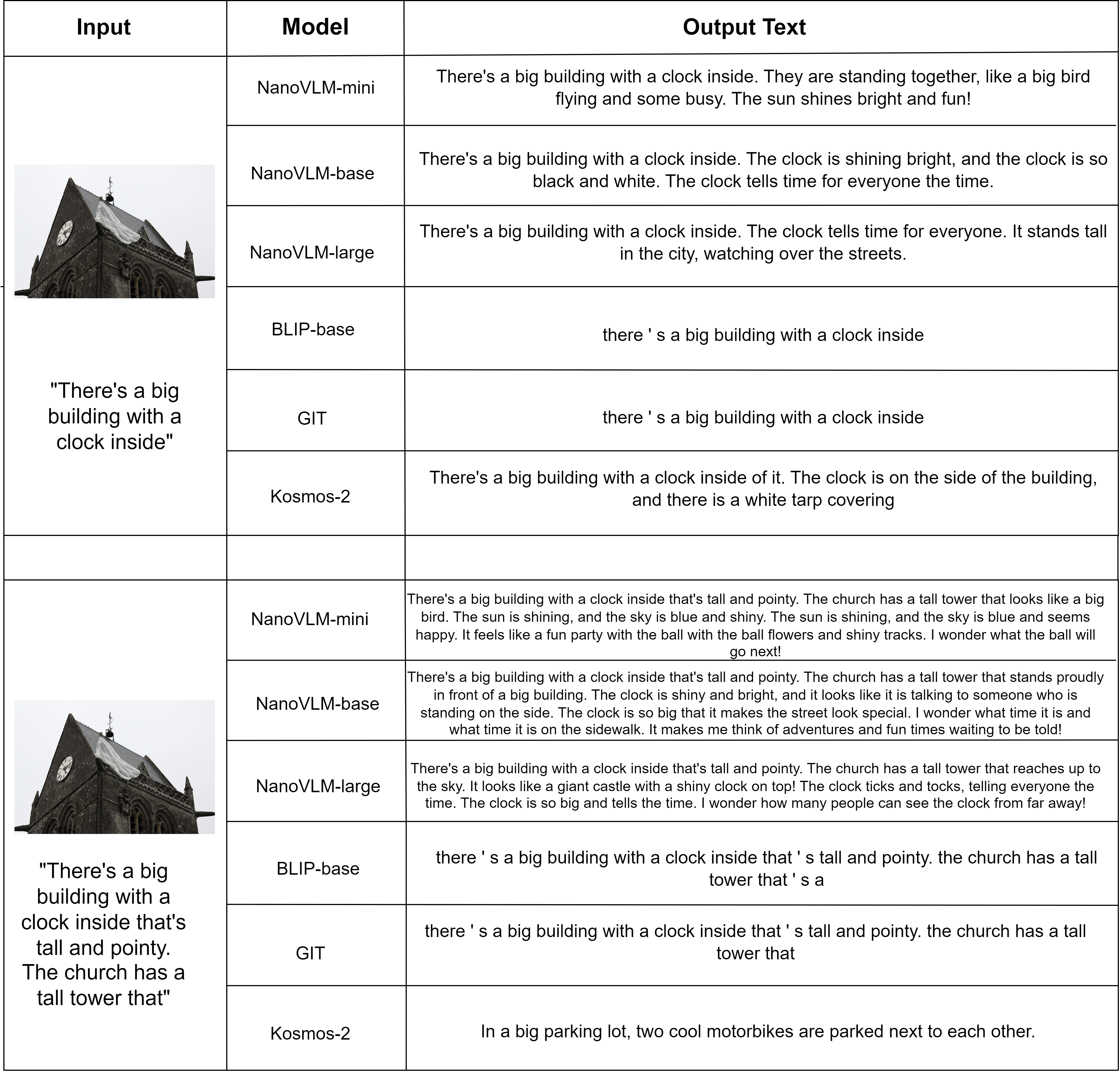}
    \caption{Qualitative comparison of image descriptions generated by NanoVLMs and other VLMs.}
    \label{fig:appendix1}
\end{figure*}

\begin{figure*}
    \centering
    \includegraphics[width=\textwidth]{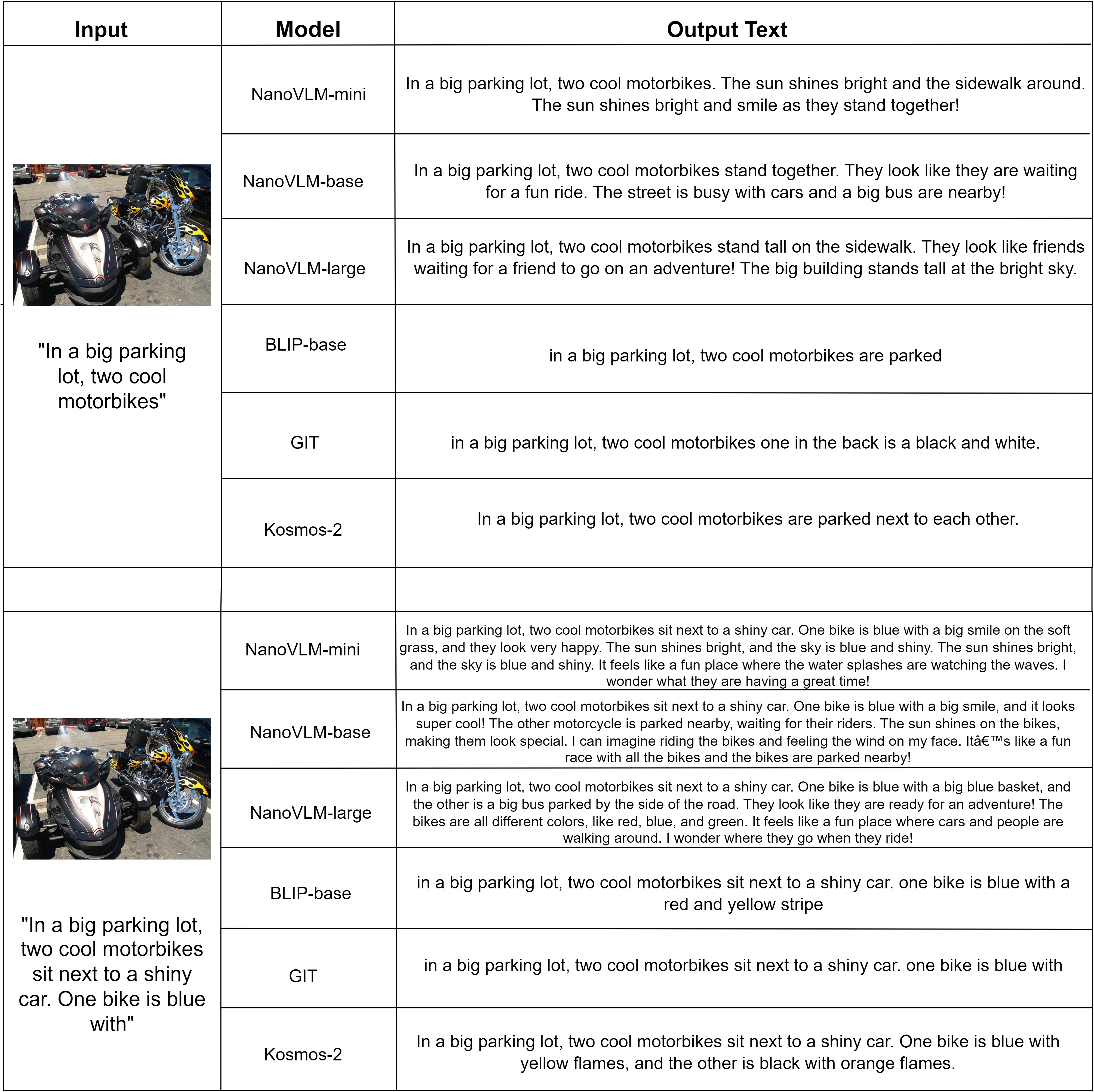}
    \caption{Qualitative comparison of image descriptions generated by NanoVLMs and other VLMs.}
    \label{fig:appendix2}
\end{figure*}

\begin{figure*}
    \centering
    \includegraphics[width=\textwidth]{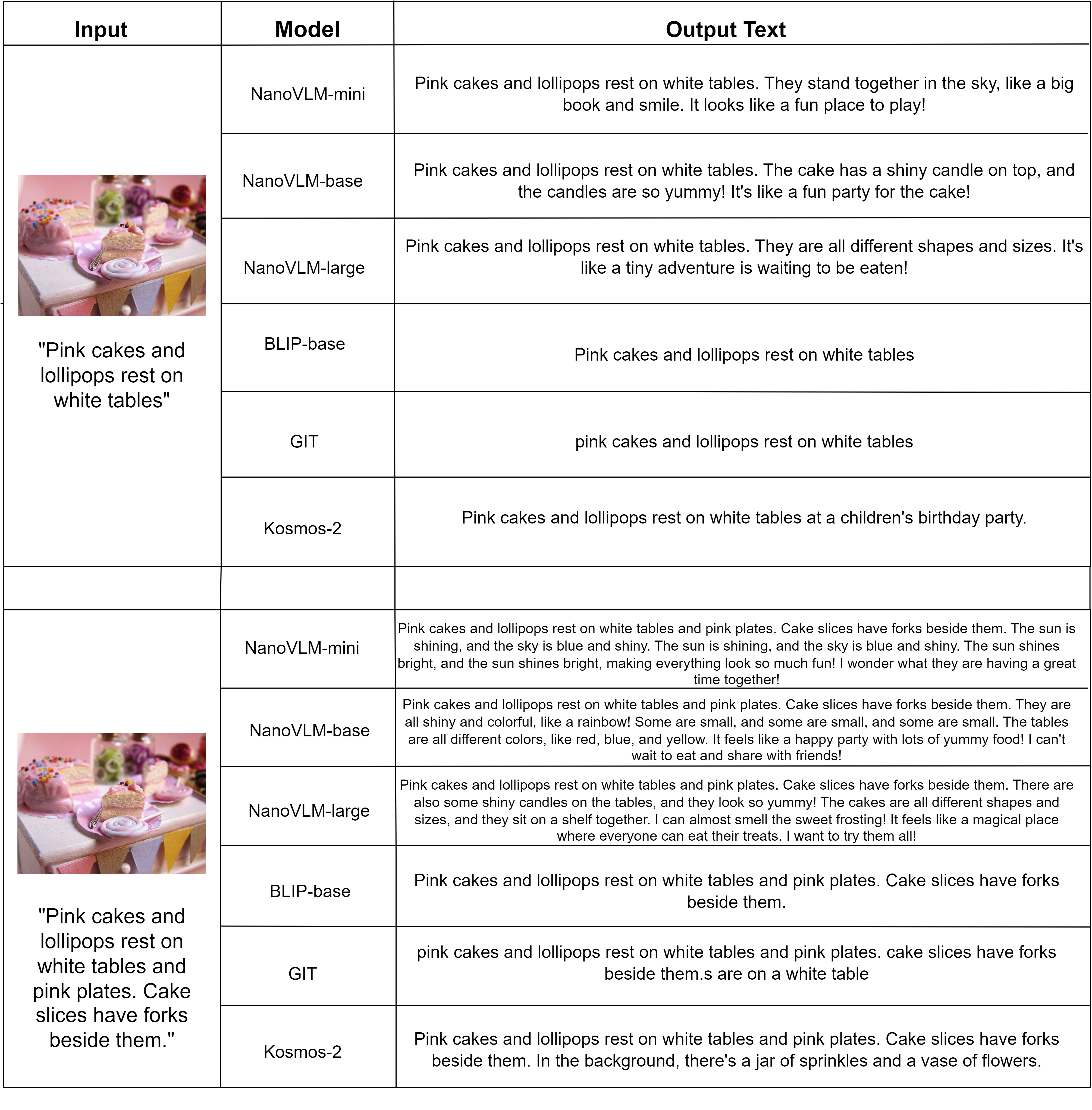}
    \caption{Qualitative comparison of image descriptions generated by NanoVLMs and other VLMs.}
    \label{fig:appendix3}
\end{figure*}

\begin{figure*}
    \centering
    \includegraphics[width=\textwidth]{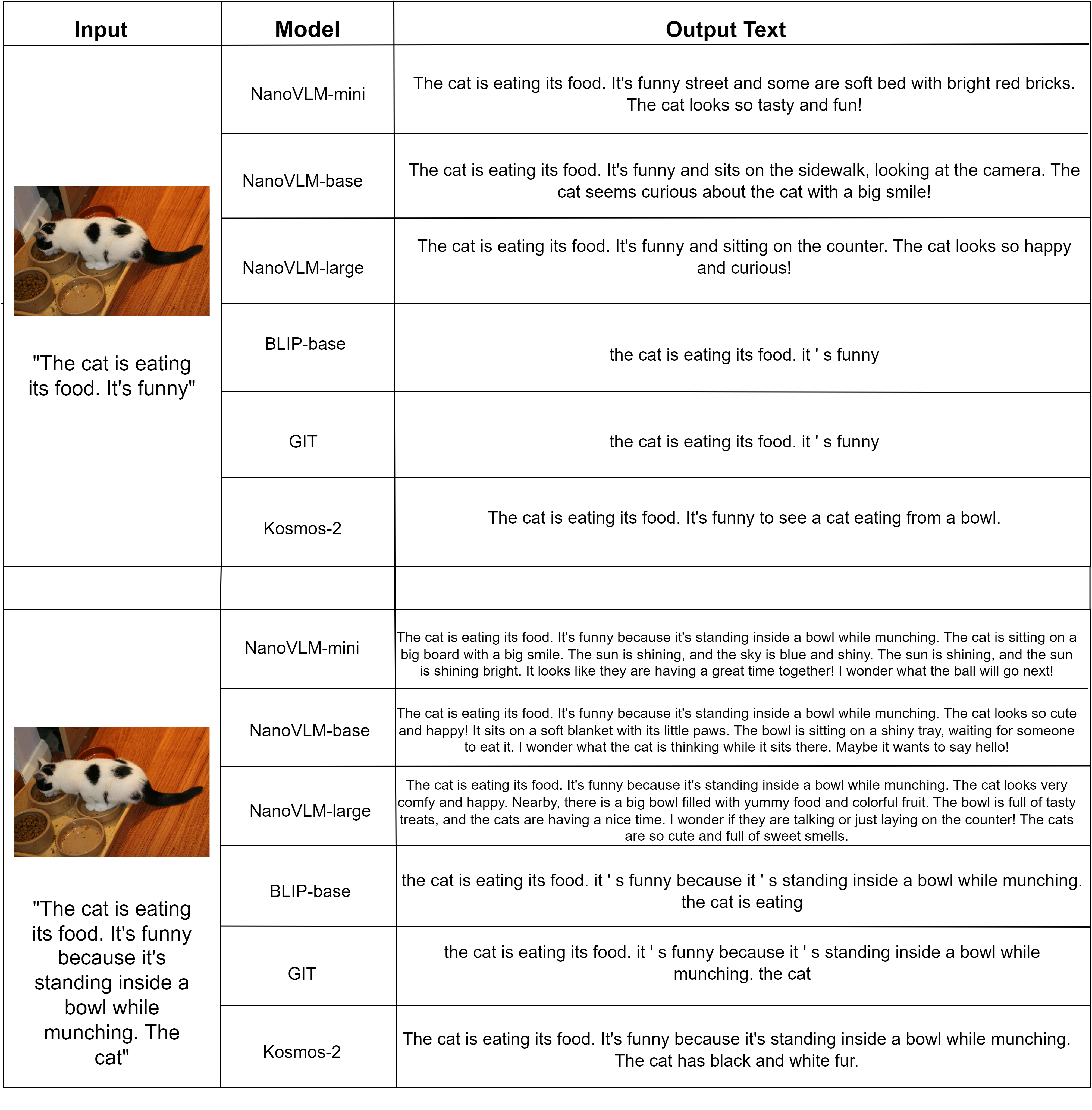}
    \caption{Qualitative comparison of image descriptions generated by NanoVLMs and other VLMs.}
    \label{fig:appendix4}
\end{figure*}

\begin{figure*}
    \centering
    \includegraphics[width=\textwidth]{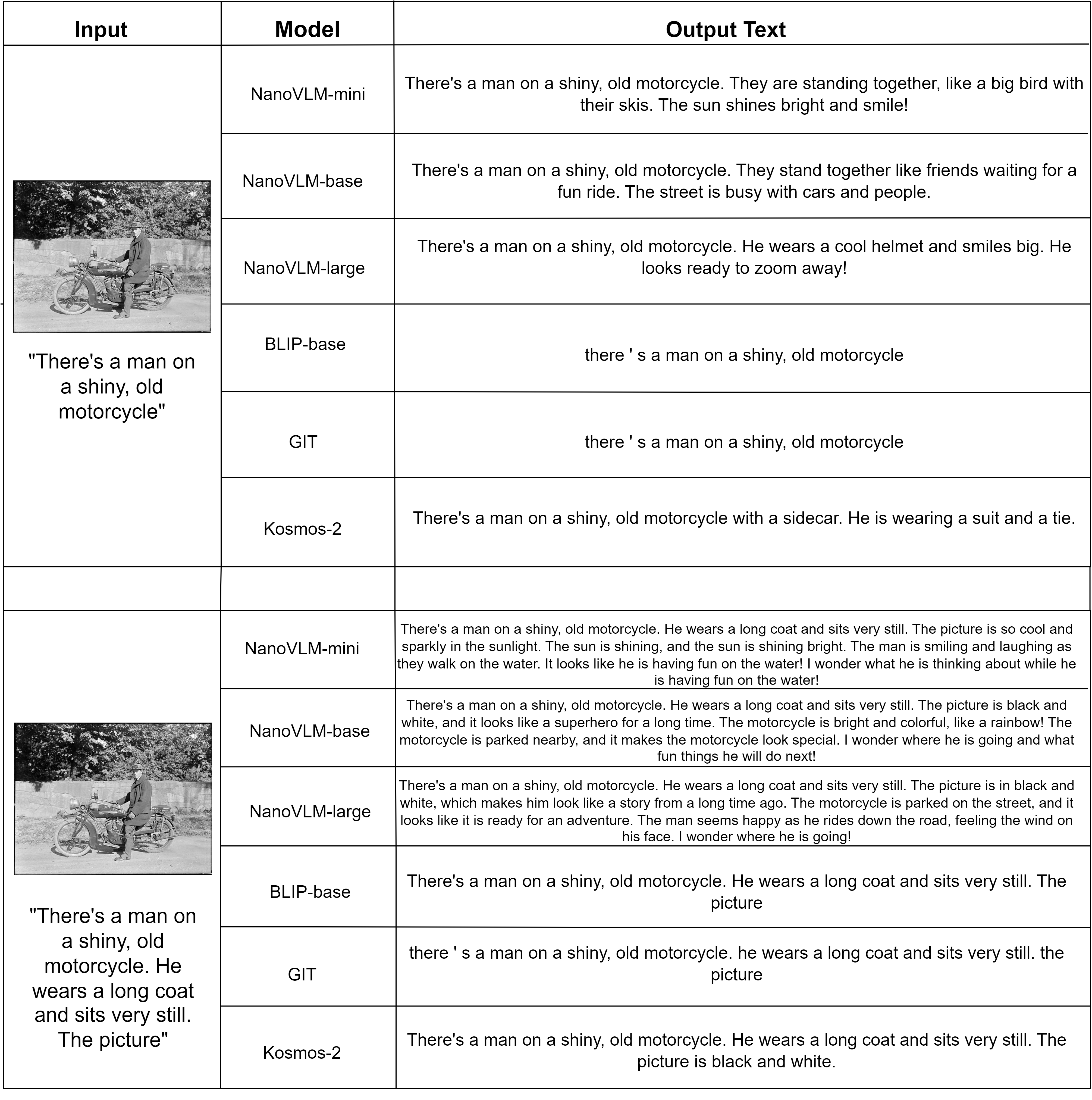}
    \caption{Qualitative comparison of image descriptions generated by NanoVLMs and other VLMs.}
    \label{fig:appendix5}
\end{figure*}

\end{document}